# Programmable Droplets as Active Matter in a Reconfigurable Environment

Ivan Shmarov†, Abhishek Sharma†, Michael Jirasek, Kristine Laws, Leroy Cronin*

School of Chemistry, The University of Glasgow, University Avenue, Glasgow G12 8QQ, UK

*Corresponding author email: Lee.Cronin@glasgow.ac.uk †Equal contribution

## Abstract

Exploring systems that bridge the gap between non-living and living matter requires a reconfigurable environment coupled to synthetic systems capable of autonomous motion, adaptive behaviour and persistent dynamics. Self-propelled droplets provide a minimal platform for such life-like behaviour, but quantitative measures capturing both persistence and adaptability to enable closed-loop reconfiguration remain lacking. Here we introduce an autonomous closed-loop robotic platform to characterise and dynamically program the autonomous behaviour of multicomponent oil-in-water droplets. We explored two chemical composition spaces, with and without external pH perturbations, analysing droplet behaviour in real time and iteratively selecting formulations. To quantify this behaviour, we introduce two descriptors, persistence and diversity. Persistence is quantified by mapping temporal dynamics in a speed–shape complexity space, where we identify a persistence domain that distinguishes active from inactive droplets. Diversity captures spontaneous or environmentally induced variation in motion and the propensity of droplets to reorient or adapt under perturbation. We show that persistence and diversity are constrained by a behavioural trade-off: high persistence limits diversity, whereas high diversity reduces persistent activity. Our results provide a quantitative framework for navigating and programming life-like behaviours in active droplets and show how closed-loop experimentation can uncover organising principles in complex active matter.

## Significance

Understanding how life-like behaviour emerges from non-living matter requires synthetic systems that sustain activity and adapt to their environment. Here we introduce an autonomous robotic platform that explores and programs the behaviour of self-propelled oil droplets, used as a minimal protocell model, by varying oil composition and environmental pH. The platform is guided by two quantitative measures extracted automatically from video: persistence, which captures sustained activity, and diversity, which describes how evenly motion spans distinct behavioural modes. Across the compositions explored, the search revealed a trade-off: high persistence limited sustained diversity. These results provide a framework for tuning life-like behaviour in reconfigurable environments and show how closed-loop experimentation can uncover organising principles in active matter.

## Introduction

Although life remains difficult to define precisely, several of its core characteristics are widely recognised[1,2]. At the level of a single individual, living systems can be distinguished from non-living systems by their ability to remain far from thermodynamic equilibrium through continuous exchange of energy and matter with the surrounding environment, and by their capacity to interact dynamically with external variations.[3–5] This exchange, captured by the concept of metabolism, sustains function by providing energy and material fluxes to maintain a non-equilibrium state[6]. Living systems do not persist passively; instead, they engage with their surroundings through dynamic behaviours, including motion and stimulus-dependent responses. Self-propulsion[7], chemotaxis[8], and phototaxis[9] are examples of such emergent behaviours arising from underlying physicochemical processes. Persistence of non-equilibrium operations and adaptive environmental interactions combined provide a significant conceptual basis for investigating life-like behaviour in artificial synthetic systems. Whilst these features describe life at the level of the individual, living systems must also be considered at the level of populations that endure across generations[10]. The long-term persistence of living species depends on reproduction and heritable variation, which together allow living systems and their functions to be maintained while enabling adaptation to dynamic environmental changes. This principle is particularly important in origin-of-life research, where the central question is how the defining principles of living systems emerged from initial abiotic systems[11]. Minimal cellular units, also called protocells, have therefore been considered as an important intermediate experimental framework for investigating the transition from non-living matter towards living systems[12,13]. Among the protocell models, self-propelled droplets are simple examples of out-of-equilibrium compartments that show sustained propulsion due to Marangoni instabilities, interfacial reaction kinetics, and spontaneous symmetry breaking[14–16]. These interactions are also regulated by droplet-droplet and droplet-environment interactions. The resulting spatiotemporal trajectories include droplet positions and shape contours on which various metrics defining droplet behaviours can be quantified. Previously, using closed-loop automated systems, artificial evolution of self-propelled droplets has been investigated using genetic algorithms towards division, motility, or vibration in multicomponent oil mixtures[17]. Furthermore, it was expanded with a configurable 3D-printed environment with increasing complexity[18] and towards open-ended exploration in the observational space defined by speed and number of droplets[19]. However, these works do not directly explore the essential feature of living systems, either biological or artificial, persistence.

Persistence can be defined as the ability to sustain and establish a set of organised processes for their continued existence against environmental perturbations. For a compartmentalised chemical system such as a self-propelled droplet as a minimalistic protocell model, persistence can be defined as its capacity to sustain its motion and features over time in the presence of environmental variations such as temperature, pH, and other chemical species in the surroundings. In general, persistence does not necessitate the protocell to remain unchanged; rather, it pertains to sustaining dynamic organisation and function over time and that characterises it as a life-like system. Here, we start by defining persistence for a self-propelled droplet on which droplet state as active (*alive*) or inactive (*dead*) at a given time can be defined and is quantifiable. We also define the diversity in the motion of the self-propelled droplets by classifying droplet trajectories into a finite set of modes of motion. By considering persistence as well as combining persistence and diversity in the droplet motion, an observational space can be defined where an autonomous research workflow can be applied to search for self-propelled droplets that exhibit life-like behaviour.

To do this work we designed and built a fully closed-loop automated experimental setup to perform the experiments with multicomponent oil-in-water self-propelled droplets, see Supplementary Media 1. The experimental arena performs dynamic pH changes by acid and base injections at the periphery of the Petri dish. The primary aim of introducing dynamic pH changes is to enhance diversity in droplet motion as an outcome of interaction with the external environment. We performed exploration and optimisation aim to direct the droplets toward a high degree of persistence and persistence-diversity in the droplet motion. We used closed-loop experimentation to perform detailed investigation to understand the dynamics of emerging multicomponent self-propelled droplets. The complete concept of persistence, persistence-diversity space, and closed-loop experimentation is shown in Figure 1.

## Speed-Shape Complexity Space and Persistence Domain

Self-propulsion and shape change are two outcomes that occur due to interfacial inhomogeneity around the droplet. The combination of these features depends on the composition of the droplet and its surroundings. Both these observational effects are considered as active behaviour and contribute to the persistence of the self-propelled droplet. In other words, any complex dynamics of a self-propelled droplet, including motion, shape change or a combination of both, can be mapped to two-dimensional space defined by speed and shape complexity. We define shape complexity as a metric that quantifies information content associated with the variations in local curvature integrated over the whole contour of the droplet. The stronger the variations in the curvature of the droplets or deviation from an ideal circular shape, the higher the information content associated with it, and contributes to the higher shape complexity. So, at each time point of droplet motion, a droplet state can be defined by its speed and shape complexity. Droplet states with low speeds and near-circular shapes are classified as inactive. Hence, in the speed-shape complexity space, thresholds can be defined on speed and shape complexity to distinguish droplet states that are inactive (*dead*) and active (*alive*). We define the domain where droplets are considered alive as the persistence domain, see Figure 1A.

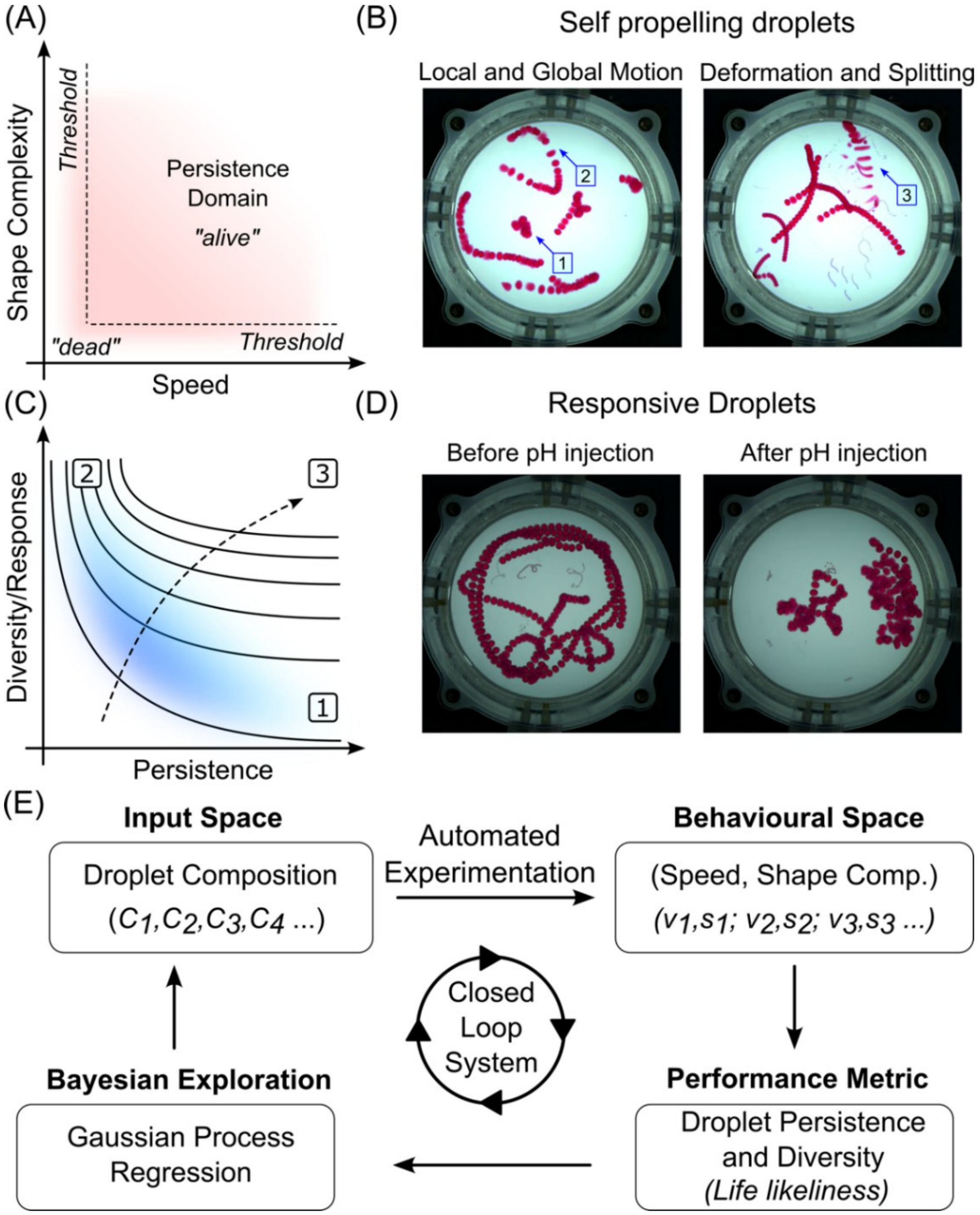

**Figure 1. Concept of Persistence and Diversity in self-propelled droplets.** (A) Pictorial representation of the persistence domain in speed-shape complexity space, where over a certain threshold self-propelled droplets are considered as “alive”. (B) Figure shows experimental overlay images of self-propelled droplets performing local and global motion and undergoing deformation and splitting, contributing to the persistence. (C) Pictorial representation of Persistence-Diversity phase space of self-propelled droplets, showing 1) high persistence and low diversity, 2) low persistence and high diversity, and 3) ideal droplets with both high persistence and diversity. (D) Figure shows experimental overlay images of self-propelled droplets in response to the injection of base, where the global dynamics of droplets become localised. (E) Closed-loop experimentation of exploration of self-propelled droplets using Bayesian Optimisation aiming for improving lifelikeness defined by high persistence and high persistence together with behavioural diversity.

As an initial test to quantify shape complexity, we used a mutual information-based formulation on a closed contour[20], which is scale-invariant and independent of rotational and translation operations. To mimic the droplet shape closed contour, we used super ellipses and Cassini Ovals to quantify the shape complexity metrics. As a first step, we use super ellipses with different degrees to define shapes with different complexities. The higher the degree $(n)$ of the super ellipse $|x/a|^n + |x/b|^n = 1$, the sharp contours represent higher information content and hence higher complexity. We estimated the mutual information on super ellipses up to degree $n = 20$ using Monte Carlo integration, which increases with the degree of the super ellipse, see Figure 2A. However, the calculation of shape complexity using mutual information is computationally expensive, and it is not practical to use on all droplet contours over the experimental timescale. So, inspired by the bending energy formulation, which is defined as a line integral of the square of the curvature along the closed path ($E_{bend} = \oint_C \kappa(s)^2 ds$), we introduced a shape metric to define the complexity of the droplet contour. Here, the integral of its squared curvature along the contour was calculated, divided by the square of the perimeter. The shape metric was defined as the logarithm of the product of that quantity and $r_{max}^3$, where $r_{max}$ is the bounding radius, which makes it less susceptible to contour size. For super ellipses, the shape metric monotonically increases with complexity (calculated using mutual information), and it is computationally inexpensive to calculate, making it ideal for closed-loop experimentation; see Figure 2A. To test the shape metric on a contour shape that mimics a droplet splitting event, we used a quartic plane curve Cassini-Oval given by $((x+a)^2 + y^2)((x-a)^2 + y^2) - b^4 = 0$. We set $b = 1$ and varied $a$ to resemble splitting $a \in [0,1]$. Inspired by the definition of bending energy, we estimated the integral of squared curvature, divided by the square of the perimeter (called BE), as well as the shape metric at different values of parameter $a$, see Figure 2B. The first metric $\log_{10}(\mathrm{BE})$ increases after splitting with strong dependence on the size. Note that the red curve considers a single contour instead of two for calculating the complexity, similar to the single contour before splitting shown in blue. The second metric (shape metric) after the splitting event falls to ca. $-0.8$, a similar value to the initial circular contour, making it not susceptible to changes in size; see Figure 2B. For additional details on the calculations on shape complexity, see Supplementary Information Section 1. At any given timepoint, as previously discussed, the state of the self-propelled droplet is defined by its speed $(s)$ and shape complexity estimated as shape metric $(m)$. In general, higher speed and complex contours signify high persistence compared to lower speed and symmetric circular shape contours. Considering this, we defined persistence of a single droplet trajectory or an ensemble of droplets $(P)$ as a summation over the whole trajectory composed of $N$ frames given by $P = \sum_{i=1}^{N} W(s_i, m_i)$, where $W(s, m)$ is the weight function. The weight function in the speed-shape metric space is defined as Equation 1,

$$W(s,m) = \min\left( g\left(\frac{s - s_0}{\bar{s} - s_0}\right), g\left(\frac{m - m_0}{\bar{m} - m_0}\right) \right) \quad (1)$$

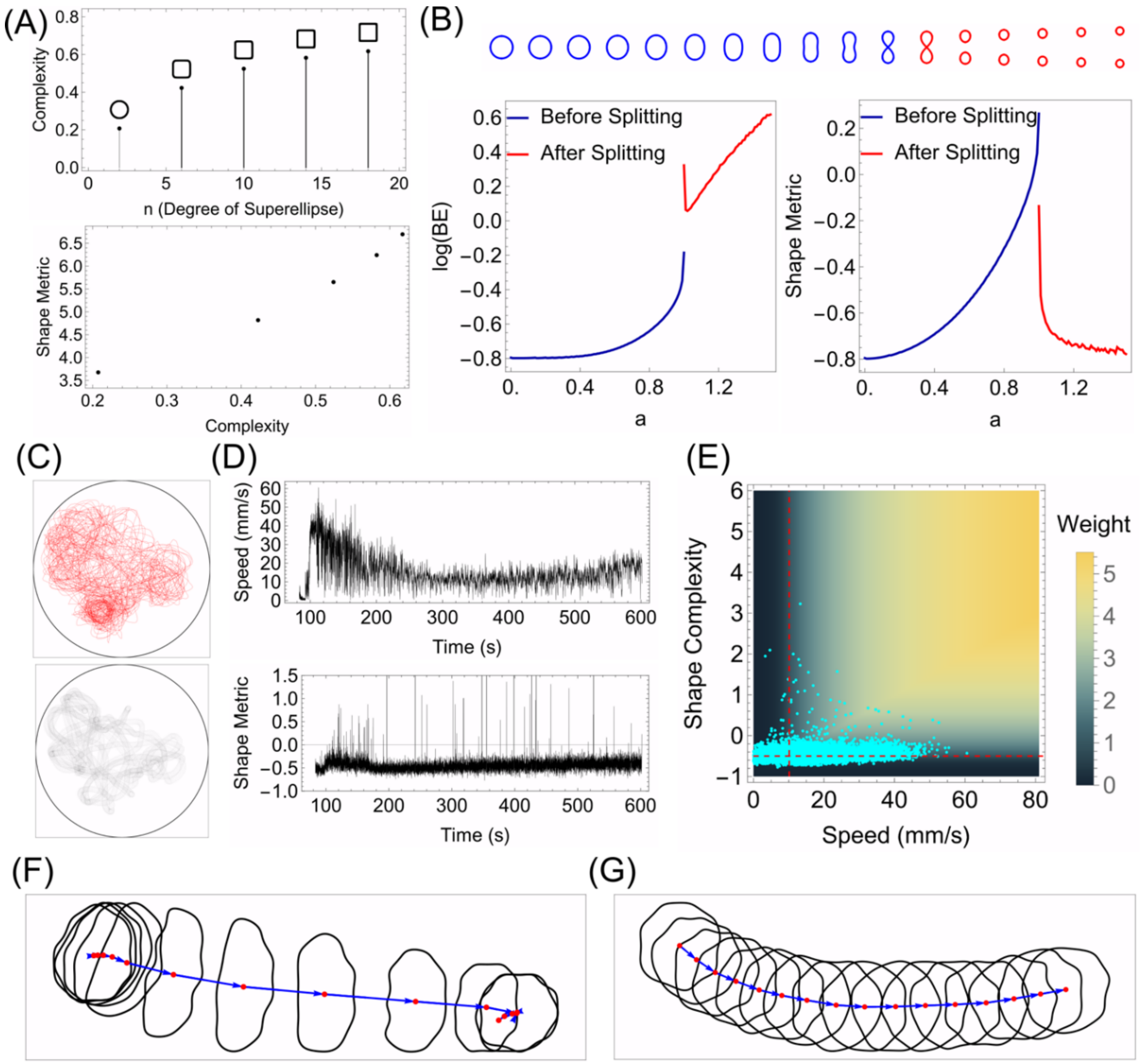

**Figure 2. Estimation of Persistence of self-propelled droplets.** (A) Top: Shape complexity of a super ellipse at different degrees, which creates sharper edges leading to higher complexity. The shape complexity is calculated based on the mutual information of the closed contours. Bottom: Estimating the complexity of a two-dimensional contour using bending energy formulation and compared to mutual information. (B) Top: Splitting event represented by Cassini Oval shape; blue represents single shape, red represents splitting into two shapes. Left: bending energy vs. $a$ parameter, keeping $b = 1$. Plot in blue represents the pre-splitting shape and red represents the bending energy after the splitting event for a single contour. Right: shape metric vs. $a$ parameter, keeping $b = 1$, showing shape metric detecting splitting event but independent of contour size. (C) Top: Example trajectory of a single self-propelled droplet in a Petri dish. Bottom: same trajectory with shape contours. (D) Speed and Shape Metric vs. time of the self-propelled droplet as shown in (C). (E) Speed-Shape Complexity space within the persistence domain. The red dashed lines represent the thresholds for distinguishing active vs. inactive droplet states. (F-G) shows droplet contours demonstrating two different modes of droplet motion: with strong changes in the curvature leading to highly erratic motion, and weaker changes leading to smoother motion.

The weight of a point in speed-shape metric space is defined as the minimum of two values, each representing a normalised coordinate processed through the kernel function $g$. The parameters $s_0$ and $m_0$ are the minimum possible values for each metric, and $\overline{s}$ and $\overline{m}$ are the domain threshold values that subdivide the persistence space into inactive or active droplet states. $g(x)$ is the normalised kernel function, which divides the persistence space into two subdomains, inactive $x \in [0,1)$ and active $x \in [1, +\infty)$. The kernel function is defined as,

$$g(x) = \begin{cases} \dfrac{e^{\alpha(x-1)} - e^{-\alpha}}{1 - e^{-\alpha}} & 0 \leq x < 1 \\ \beta \log_{10}(x) + 1 & 1 \leq x \end{cases} \tag{2}$$

with $\alpha$ = 4 and $\beta$ = 5, where the two branches meet continuously at $x = 1$. For more details on calculation of persistence calculations and details of various parameters, see Supplementary Information Section 1.

As an example, we consider a single self-propelled droplet trajectory for calculating persistence; see Figure 2C for the position and shape contours of the droplet over the experimental period. The speed and shape metrics are calculated as described previously and are shown in Figure 2D. Each timepoint of the droplet state defined by speed and shape complexity is shown within the persistence domain (in cyan), see Figure 2E. The background density plot represents the weight function over the persistence domain. The points at the left and bottom of the dashed lines (in red) are inactive droplet states, and the top and right correspond to active droplet states. The persistence is calculated by summing the time points of the droplet trajectory on the weighted function ca. 14780. The calculation of persistence on the droplet trajectories is described in detail in the Supplementary Information Section 1.

## Experimental Setup

Exploring persistence and the trade-off between persistence and diversity required hundreds of oil-in-water droplet experiments over a continuous input space of four different reagents (oils). Such a space cannot be explored manually: every experiment must be prepared, performed, and reset identically for the results to be comparable and exact for the closed-loop experimentation. Automation resolves only the throughput half of the problem, the oil mixture space is too large for exhaustive enumeration, making the selection of the next experiment equally limiting. Closing the loop therefore requires each result to be analysed on acquisition and fed back into a model that proposes the next recipe. We therefore built a platform that runs the complete cycle without human intervention: it prepares a recipe (combinatorial mixture of oils), performs the experiment, analyses the recorded videos, scores the result, and chooses the next recipe from the machine learning model based on the data collected so far, see Figures 1E and 3.

We chose a pipetting robot (Opentrons OT-2) for the liquid handling operations because it dispenses volumes in the microlitre range precisely. Its deck has multiple slots which are customisable, so we could perform the entire experiment within the robot, see Supplementary Media 1. We mixed the oils in vials on an orbital shaker on the deck, which homogenised the mixture before it is dispensed. The experimental arena is a glass Petri dish, which does not degrade over multiple experiments and retains no residue between experiments. A 3D-printed holder fixed the dish in one of the slots and carries the injection ports at its rim, so that the pipette places droplets at programmed coordinates, see Figure 3A. The injection ports deliver the basic and acidic solutions at four fixed positions around the dish, such that localised pH changes in the aqueous phases can be introduced, creating spatiotemporal gradients. A second 3D-printed unit, driven by a stepper motor, is mounted beside the dish and lowers its tubes into the arena when needed, see Figure 3B. One tube reaches the bottom of the dish to draw the contents to waste, while the other, held just above the bottom, delivers rinse water and fresh aqueous phase. Both lines are served by syringe pumps, such that the volumes delivered and removed are the same in every experiment.

We record the motion of droplets from above with a camera served by a dedicated Raspberry Pi computer, which transfers the finished recording to the host PC that runs the experimental loop. The whole arena lies within the field of view, and at 30 frames per second a droplet moves only a fraction of its own width between frames, so individual droplets can be followed from frame to frame. Each recording runs for 17 or 30 minutes, depending on the experimental run. A complete iteration of the experimental run is shown in Figure 3C. The main PC suggests an oil mixture recipe, the OT-2 prepares that mixture, and the unit cleans the arena and refills it with fresh aqueous phase pre-mixed with surfactant. The pipette then dispenses the droplets and the recording begins. Once the recording is complete, it is analysed and the recipe is scored. The recipe and its score are added to the dataset, the model is updated, and the next recipe is generated. One complete experiment cycle takes 40 to 60 minutes, and a full rack of 60 mixing vials (on the orbital shaker) lasts around two days. The hardware and the control software are described in Methods and in Supplementary Information Section 2. See Supplementary Media 1 for various preparation steps required for experimentation.

Within the loop, once the experimental video is recorded and transferred to the host PC, the video was analysed framewise, and after performing image segmentation, droplet contours were extracted. We performed Elliptical-Fourier analysis of the contours, and the complex droplet shapes were fitted up to 10 harmonics[21]. The shape complexity, estimated as a shape metric (modified formulation from bending energy), was calculated from the shapes.

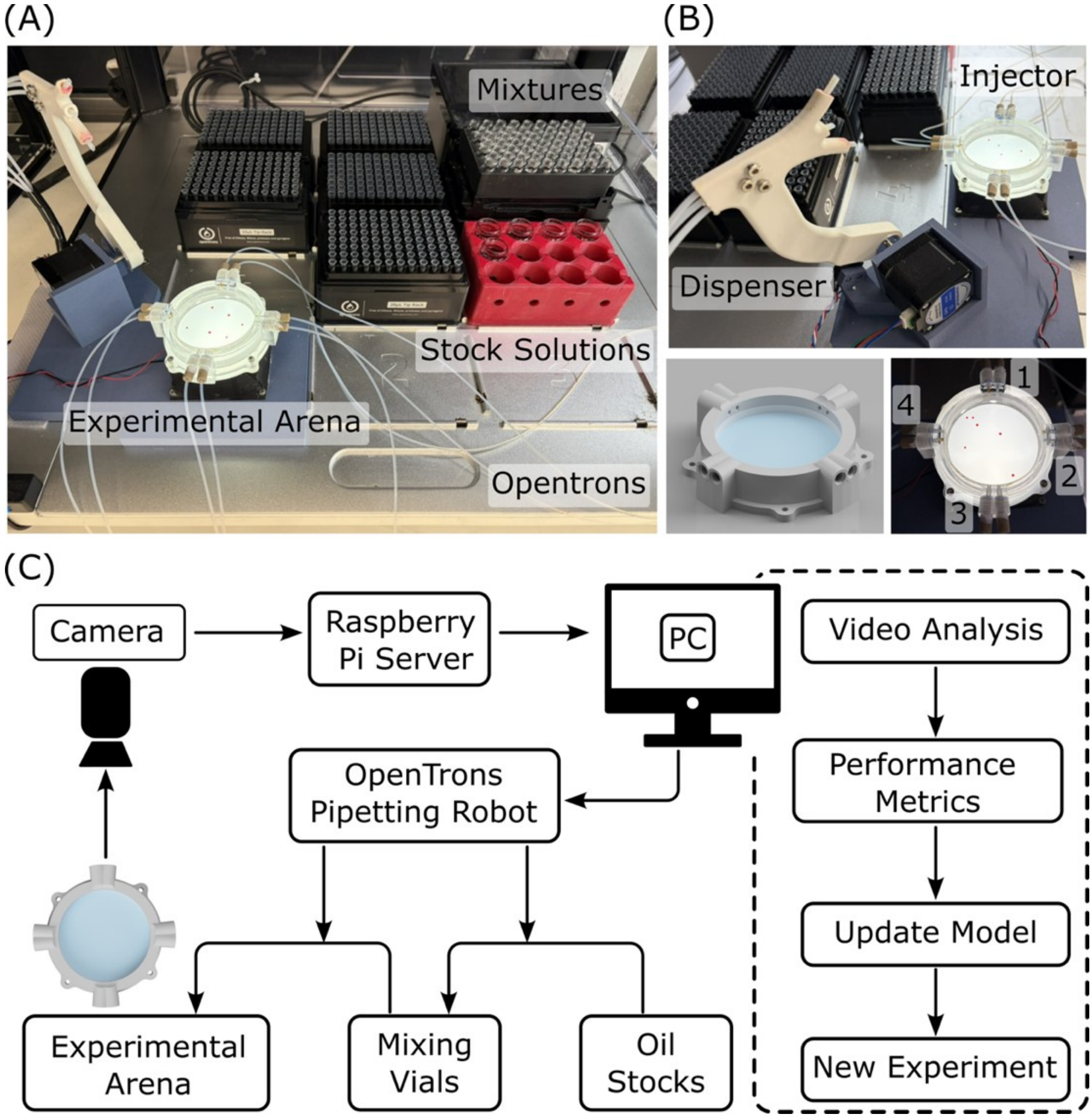

**Figure 3. Experimental setup for closed-loop experimentation of self-propelled droplets.** (A) Image shows experimental arena using Opentrons (OT2) pipetting robot with Petri dish fitted within 3D-printed holder with external injectors. (B) Image shows dispenser unit for delivering continuum phase (aqueous) and cleaning Petri dish between consecutive experiments, rendered image 3D-printed arena to hold injectors, top view of assembled experimental arena with self-propelled droplets (in red) with four injectors. (C) Complete experimental strategy for closed-loop experimentation with robotic experiments performed with customised Opentrons (OT2) setup, imaging using Raspberry Pi HQ camera, and video analysis, and Bayesian optimisation on the host PC, see Supplementary Media 1.

The droplet contours from consecutive frames were stitched together to create the trajectories on which droplet speed can be calculated. The analysis was performed for each detected droplet and using the estimated droplet states defined by speed and shape complexity, the persistence metric over all the trajectories was calculated as an optimisation target in closed-loop experimentation.

## Exploring self-propelled droplets towards high degree of persistence

As a first closed-loop experiment, we considered a four-dimensional input space comprising oils: pentanol (Pent), octanol (Oct), diethyl phthalate (DEP), and octanoic acid (OA)[19]. The aqueous phase contained surfactant tetradecyltrimethylammonium bromide (TTAB) and the pH of the solution was adjusted to 10 by adding NaOH. A basic medium promotes the hydrolysis of DEP, which can make interesting behavioural dynamics of droplets with dynamic speed and shape changes. The aim is to find the oil mixture of the four components, which gives the maximum persistence score, suggesting the highest life-like behaviour. Using the experimental design based on the setup described in the previous section, oil-in-water droplet experiments were performed in a closed-loop using Bayesian Optimisation on a surrogate model defined by Gaussian process, see Figure 4, Methods and Supplementary Information Section 4 for details. In the closed-loop approach, the initial oil combination recipes were sampled from the Dirichlet distribution on a 3-D simplex.

Further experiments explored toward a high degree of persistence were sampled by the implemented Bayesian Optimiser, and the details are described in the Supplementary Information Section 4 and Figure 4A shows the persistence score as a function of the iteration number in the loop. The vertical blue line marks the cutoff between the initial seed experiments and the model-suggested oil mixtures. The dashed red line shows the rolling best performance of oil mixtures, showing the gradual rise in persistence score, demonstrating the efficacy of the implemented closed-loop optimisation. Some experiments scored lower in persistence than earlier ones, which is expected. Each new oil mixture was suggested according to one of three strategies, between which the loop alternated. A strategy is a heuristic that uses the recipes and persistence scores collected so far to suggest the next oil mixture to test. Only one of the three was exploitative, suggesting the recipe with the highest predicted persistence. The other two were exploratory, so the loop did not aim at a higher score at every experiment. The strategies are described in full in Supplementary Information Section 4.

The highest persistence observed at the end of the Bayesian Optimisation loop was $2.65 \times 10^5$ after 60 iterations. The volumes of various oil components in the mixtures are: $\mathrm{Pent} = 0.12$, $\mathrm{Oct} = 0.29$, $\mathrm{DEP} = 0.57$, $\mathrm{OA} = 0.005$. Figure 4B-E shows details of the self-propelled droplets experiment with the highest observed persistence. The droplet ensemble distribution in the persistence space shows a high density of droplet states with speed ca. 15 mm.s$^{-1}$ (shown in yellow), which contributes to high observed persistence, see Figure 4B. See Supplementary Information Section 4 for distributions in persistence space at different experimental iterations for comparison. To investigate it further, we calculated the average droplet speed of all the droplets as a function of time over the elapsed experiment time. We observe different modes of motion over the course of the experiment, where the average speed falls initially, then rises again, and falls till the end of the experiment; see Figure 4C. To observe this behaviour in detail, we extracted six droplet trajectories starting at different times in the experiment, highlighted as a-f in Figure 4C.

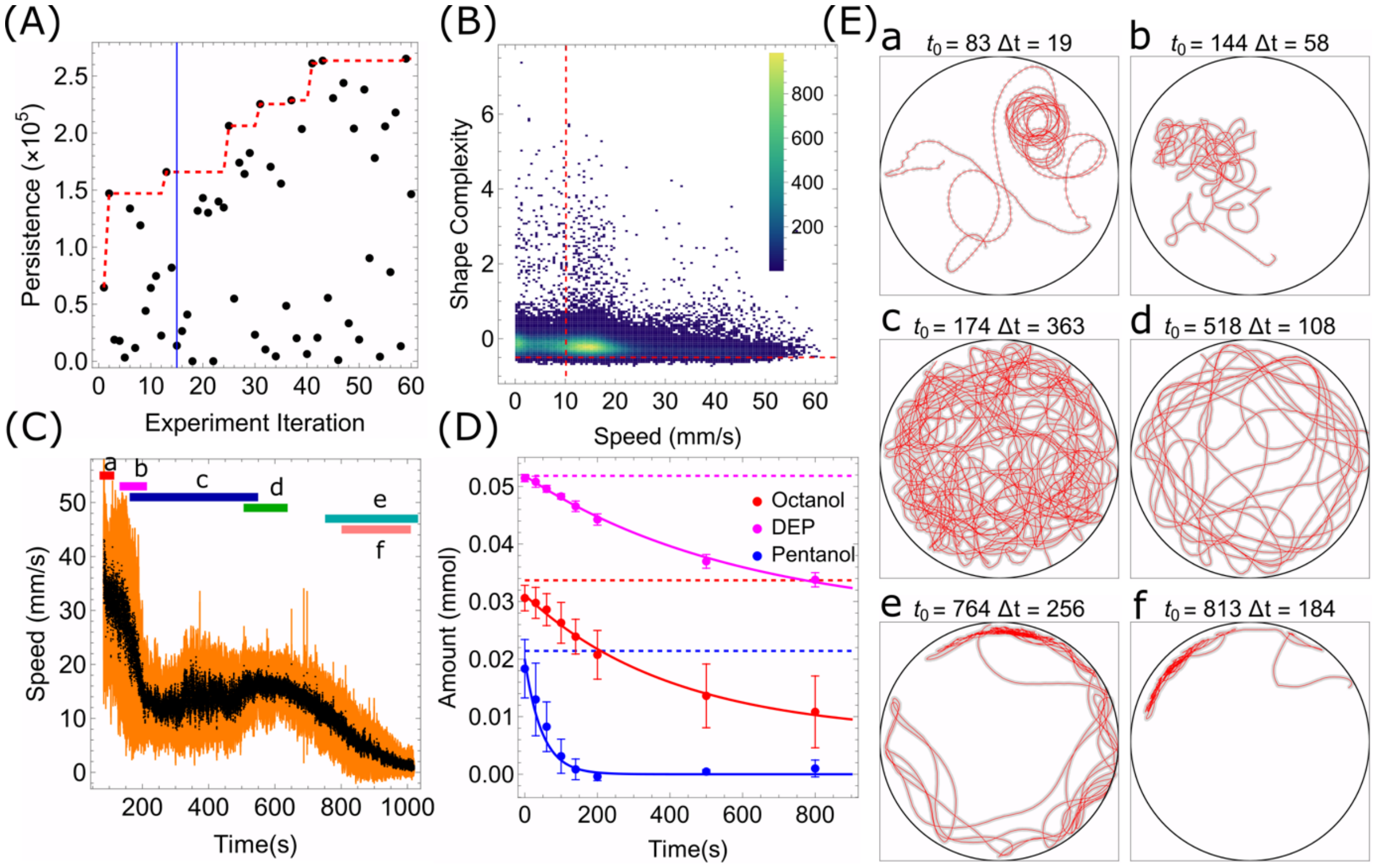

**Figure 4. Bayesian Optimisation of self-propelled droplets towards high persistence.** (A) Optimisation of persistence over the experimental iteration using closed-loop experimentation. (B) Density distribution of droplet trajectories in speed-shape complexity phase space of the experiment with highest persistence. (C) Average speed (and standard deviation) of self-propelled droplets over the experimental time with six different trajectories highlighted (a-f). (D) Amount of oils in droplet phase (Octanol, DEP, and Pentanol) over time estimated using NMR experiments. (E) Six different trajectories (a-f) sampled at different times (as shown in C) over the course of the experiment. These include local and global modes of motion over the course of 900 seconds.

The droplet trajectories (a-f) are shown in Figure 4E, where distinct modes of motion are visible. At the initial phase, the droplet motion is highly localised, rotating in circles as shown in Figure 4E(a). After that, the motion is still localised but with sharper changes in the directionality of the trajectory, see Figure 4E(b), before transitioning to global motion for an elongated period, which contributes towards high persistence, see Figure 4E(c-d). Eventually, the droplets start slowing down, and their motion becomes confined to the periphery of the Petri dish, see Figure 4E(e-f). See Supplementary Media 2 for acquired experimental video of self-propelling droplets at highest persistence. To investigate further the temporal dynamics of droplets, we performed NMR measurements of the continuum phase sampling at different times to estimate the amount of oils in the aqueous phase. With this, the time-dependent oil dissolution was estimated; see Figure 4D. At the initial phase, pentanol dissolved fastest, driving the initial dynamics of localised motion ($t_{1/2} \approx 32$s), with octanol dissolving gradually over the course of the experiment. DEP ($t_{1/2} \approx 414$s) together with octanol ($t_{1/2} \approx 284$s) dissolution drives the later dynamics of the droplets. We attribute this to hydrolysis of DEP, forming ethanol which leads to fast dissolution into the aqueous phase. See Supplementary Information Section 5 for more details.

## Interaction with dynamic pH changes and diverse modes of motion

As a next step, we introduce external perturbations into the pH of the continuum aqueous phase to interact with the dynamics of droplets. Here, we use a different combinatorial mixture for the droplet oil phase: ethyl salicylate (ES), hydrocinnamaldehyde (HCA), 2-hexyldecanoic acid (HDA), and di(2-ethylhexyl)phosphoric acid (DEHPA). This combinatorial mixture was selected to have strong self-propulsion due to ES[22] together with HCA and interaction with the surrounding pH due to DEHPA and HDA. As part of the experimental design, we used four acid-base injection pairs placed at the periphery of the Petri dish as shown in Figure 3B. We used HCl ($\mathrm{pH} = 1$) and NaOH ($\mathrm{pH} = 13$) as acid and base to create dynamic pH changes in the Petri dish via syringe pumps. As an example, to test pH changes, we injected a fixed volume of acid and base in the aqueous phase with universal indicator in the clockwise order. The temporal domains of pH changes: basic (in blue), acidic (in orange/yellow), and original domains close to neutral pH (green) are shown in Figure 5A, see Supplementary Media 3. The red dots indicate the position of the last active injector that performed the injection operation. We explored the interaction of pH changes with the oil mixtures, and we observed that droplet dynamics interact strongly with basic conditions, creating novel behaviour, but not with strongly acidic conditions. As an example, we compare the dynamics of self-propelled droplets in the presence and absence of pH changes by base injections. In the absence of pH changes, the average droplet speed and behaviour remain consistent over the course of the experiment. With the base injections, strong variations in the average droplet speed and behaviour occur; see Figure 5B. The difference in the dynamics of droplets in the absence and presence of pH gradients is clearly visible within the distribution in the persistence domain, see Figure 5C.

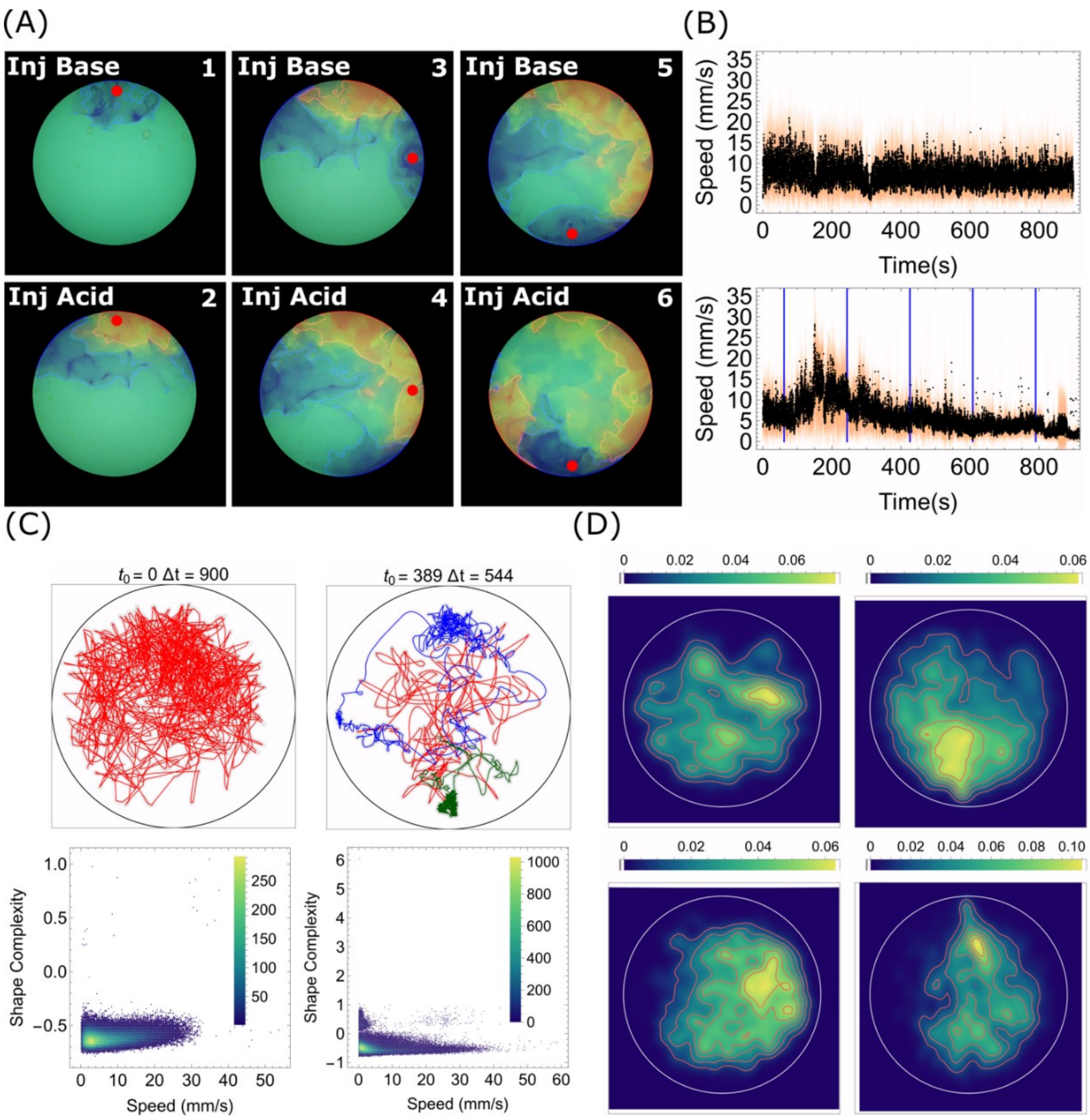

**Figure 5. External perturbations to droplet trajectory via acid/base injections.** (A) Figures show pH changes over the Petri dish with simultaneous acid and base injections. The red/yellow and blue regions show acidic regions and basic regions. The red point shows the point of injection. (B) Average speed (and standard deviation) of self-propelled droplets over the experimental time in the absence and presence of base injections. The vertical lines in blue show the time of base injections. (C) Persistence domain and trajectory of a self-propelled droplet in the absence (left) and presence (right) of base injections. In the presence of the injections, the trajectory is shown in three different colours: red: before the first injection, blue: after the first injection, green: after the second injection. (D) Density distribution of droplets over consecutive injections. Top Left: before the first base injection, Top Right: after first injection (base only), Bottom Left: after second injection (base only), Bottom Right: after third injection (base only).

In the absence of pH changes, both speed and shape complexity of the droplets are relatively homogeneous, as seen in the persistence domain, and a single trajectory plotted over the whole experimental time. In the presence of pH changes, the droplet motion becomes erratic, and the variations in both speed and shape complexity are much stronger, as seen by the distribution of droplet states in the persistence domain. An example trajectory shows that the droplet motion switches from global to local when experiencing high basic domains from the injection operations. The droplet gets trapped locally with strong vibration and stronger shape changes. A single trajectory in Figure 5C shows locally trapped states at different injections, particularly shown in blue and green. The effect of the base injections can be seen by looking at the density of droplet positions between each injection period for each injector, see Figure 5D.

These experiments demonstrate that the motion of self-propelled droplets has two key aspects of systems with life-like behaviour: persistence and diversity. Persistence in general represents the motility of the droplet and its capacity to stay in an active state, and diversity represents its response to the change in the local environment. In general, the response emerges as an outcome of sensing the local environment and actuating accordingly based on the internal state of the droplet. Some of the behaviours that emerge as an outcome of the response are chemotaxis in droplets[23,24], collective rotation of clusters[25], and chemical echolocation to escape obstacles[26], among others. To quantify the response as an outcome of the internal dynamics of the droplet or interaction with the environment, we define a metric to quantify the diversity in the droplet motion over the experimental timescale. For simplicity, here we limit the definition of diversity to individual droplet trajectories, not the collective motion, to keep it as an intensive property that does not depend on the size of the droplet ensemble. This becomes a complementary property to persistence; usually, high-persistence droplets do not respond well to environmental changes. So, combining persistence and diversity into an observational space enables search for a trade-off between persistence and diversity, and optimal droplets should possess a good combination of both features. The diversity within a droplet trajectory can be characterised by two complementary metrics describing the extent and direction of motion. The extent is defined as the maximum displacement within a prescribed time window and quantifies the spatial scale explored by the droplet, distinguishing confined motion from trajectories spanning the arena. Unlike trajectory arclength, it is insensitive to local path tortuosity. The direction of motion is characterised by the turning rate, which quantifies temporal changes in trajectory orientation. By applying a threshold on these metrics, the droplet's instantaneous motion can be classified into one of the four modes of motion: Local Linear (LL), Global Linear (LG), Local Rotational (RL), and Global Rotational (RG).

As an example, we chose a single droplet trajectory from an experiment lasting for up to ca. 200 seconds; see Figure 6. The droplet performs a variety of motions with localised motion initially, which later becomes global. We classified the trajectory into these four modes, see Figure 6A-B. Individual modes of motion are shown in Figure 6C-F, showing local and global linear modes which can be easily distinguished based on the density of the points along the trajectory. Similarly, the local and global rotational modes can be distinguished based on the local radius of curvature and density of the points along the trajectory. Over the whole course of the experiment, the droplet undergoes extensive mode switching which can be seen in the temporal profile, see Figure 6G. For each trajectory, the distribution of motion modes is obtained from the fraction of time points assigned to each mode. Aggregating these distributions across all recorded trajectories yields the overall mode distribution for the experiment. We define the diversity of motion modes as the entropy of this aggregate distribution. The complete mathematical formulation used to quantify mode diversity across the droplet ensemble is provided in Supplementary Information Section 1.

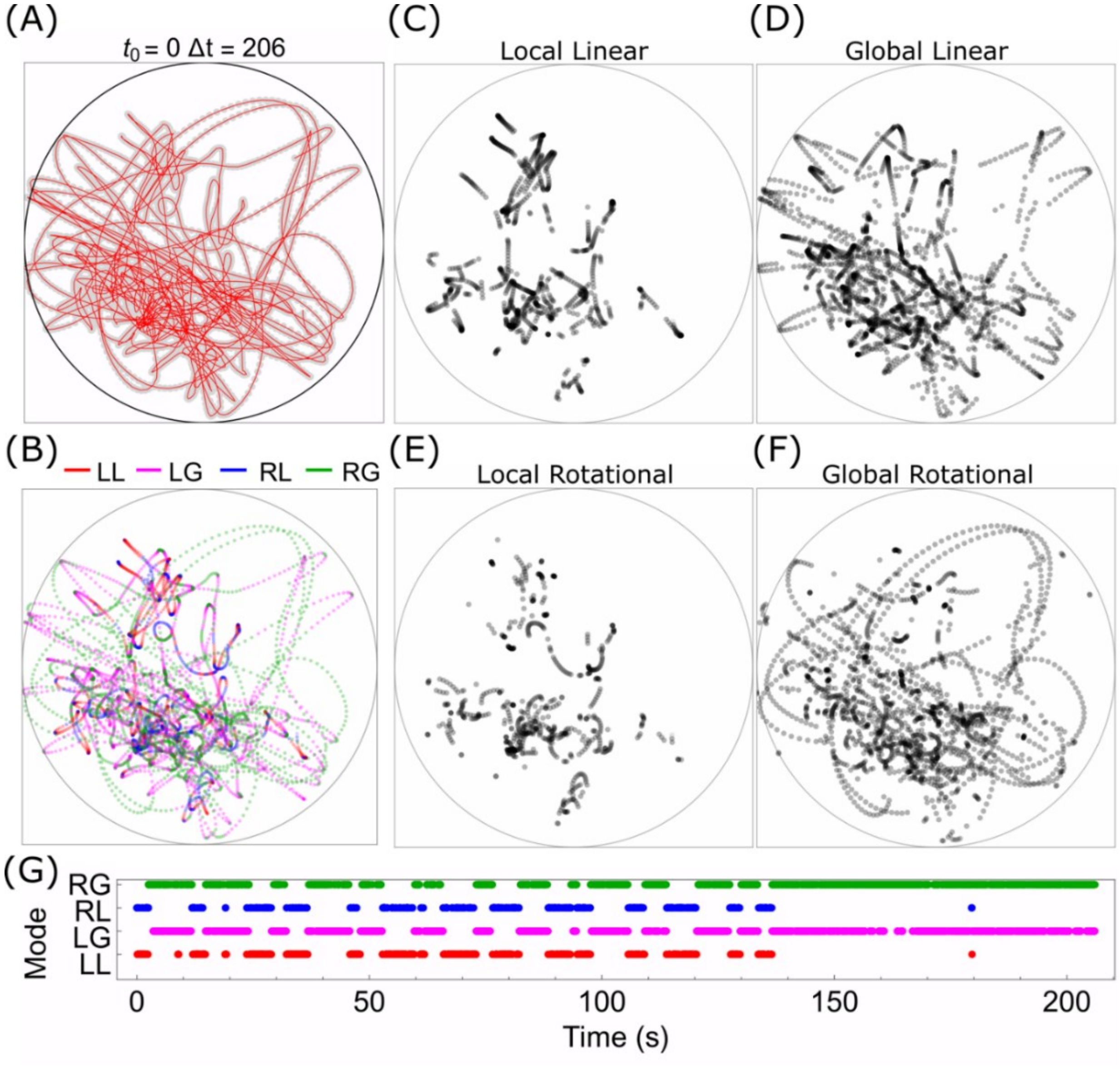

**Figure 6. Different modes of motion of self-propelled droplets.** (A) Example of a characteristic trajectory of a self-propelled droplet observed at the start of the experimental measurement and lasting for 206 seconds. The grey points are the sampled points, and red line is an interpolated function. (B) Splitting the trajectory shown in (A) into four different modes: Local Linear (LL, shown in red), Global Linear (LG, shown in magenta), Local Rotational (RL, shown in blue), and Global Rotational (RG, shown in green). (C–F) Individual trajectories of all four motions shown in (B) for clarity. (G) Temporal changes of the trajectory over the different modes of motion. The trajectory shows localised dynamics together with global motion up to ca. 130s and then turns into dominantly global motion.

## Exploring self-propelled droplets in persistence-diversity space

As discussed in the previous sections, the combination of persistence and diversity of motion modes could be used together to explore the self-propelled droplets with a trade-off between motility by persistence and response to the external environment by mode diversity. So, we define the optimisation target as $T = P \times S^{\alpha}$, where we chose $\alpha = 10$ to have a good balance between persistence and diversity, and not to only direct the goal towards high persistence. In principle, the parameter $\alpha$ can be chosen to balance persistence and diversity in the exploration loop. In this experimental loop, we used the combination of four oils: ES, HCA, HDA, and DEHPA for the droplet phase, and the surfactant SDS in the aqueous phase. The injectors were active with base injections every 180 seconds; see Methods for details. Similar to the previous run, the optimisation target showed an iterative increase over the course of ca. 120 experiments, see Figure 7A. The initial 32 experiments were seed experiments, shown to the left of the blue line. To check sampling of experimental data in the observational space, we plotted persistence vs. diversity over the course of experiments, see Figure 7B. The plot clearly demonstrates that at higher persistence values, $P > 1.5 \times 10^5$, high diversity of the motion modes cannot be sustained, suggesting a trade-off between persistence and diversity. The maximum diversity was observed around $P \sim 1 \times 10^5$, and then it decreased as persistence rose further, so beyond that point the trade-off set an upper bound on the diversity that could be sustained. Over the course of experiments, although it is expected that persistence does not increase steadily, similar to Figure 4A, it still approaches a convergence, as evidenced by the decreasing spread of values over time, see Figure 7C.

To better illustrate the trade-offs between persistence and diversity as an outcome of the optimisation loop, Figure 7D-L shows examples from the exploration run with different characteristics: highest diversity ($S$), highest persistence ($P$), and highest persistence-diversity function ($T$). The first droplet ensemble with oil ratios ($\mathrm{ES} = 0.66$, $\mathrm{HDA} = 0.03$, $\mathrm{HCA} = 0.18$, $\mathrm{DEHPA} = 0.13$), which shows highest diversity with a characteristic trajectory ($\Delta t > 900s$), persistence domain, and mode diversity is shown in Figures 7D-F. The overall target score is low with $T = 2.25 \times 10^6$, $P = 87438$, $S = 1.38$. Here, the persistence space shows a high distribution of droplet states at lower speeds, indicating low persistence; however, the whole trajectory shows approximately equal distribution among the four modes of motion, leading to high diversity. The key contributor to the high diversity is DEHPA, which interacts strongly with the base injections, transitioning the motion from global to local such that the motion modes keep switching between the four modes, up to 800 seconds, after which the droplet eventually slows down, switching to local linear and rotational modes.

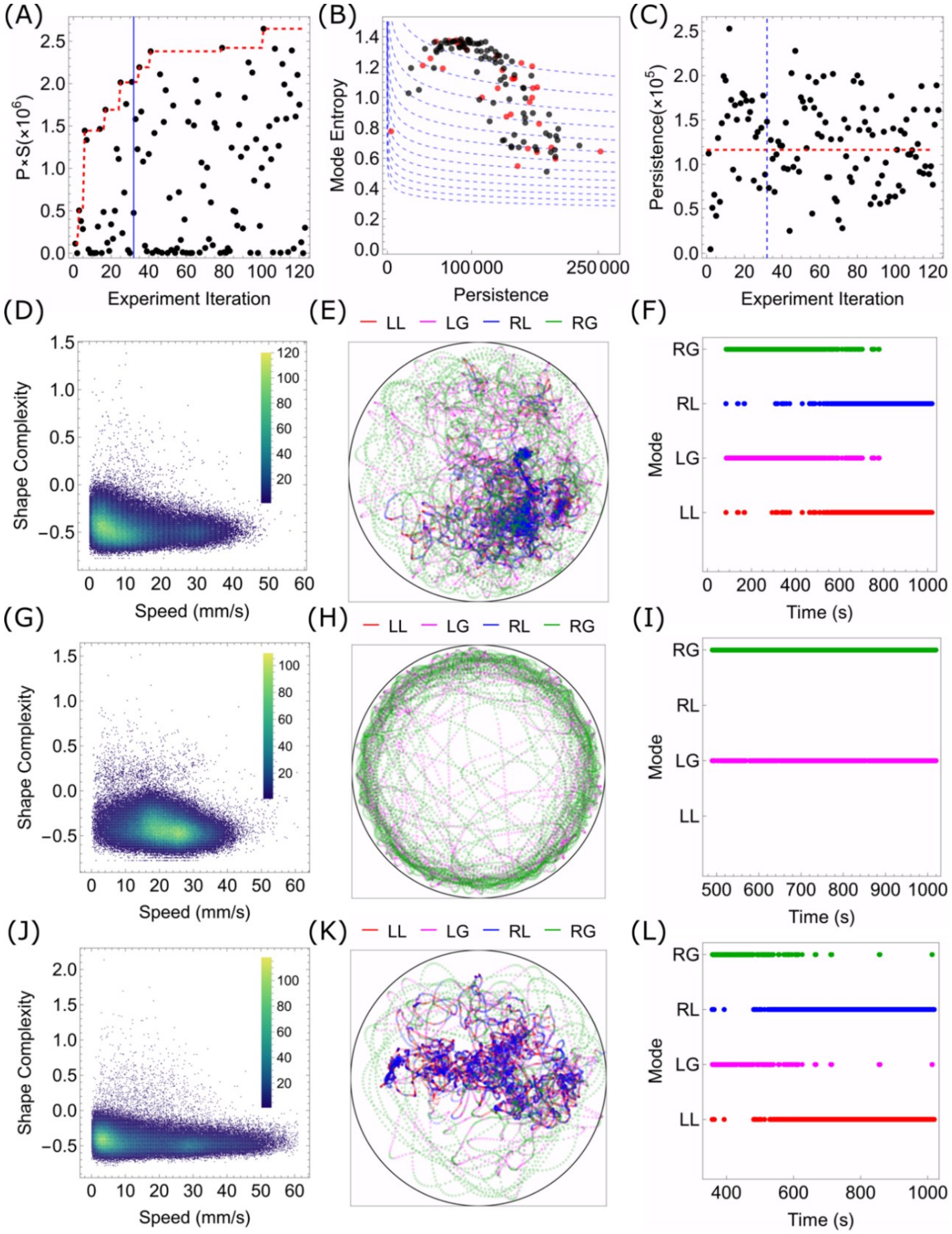

**Figure 7. Bayesian Optimisation of self-propelled droplets towards high persistence-diversity behaviour.** (A) Optimisation of persistence-diversity function ($T = P \times S^{10}$) over the experimental iteration using closed-loop experimentation. The points along the dashed red line represent the highest performing individuals and the points to the left of the blue line represent initial experimental dataset. (B) Distribution of the experimental points in two-dimensional space represented by Persistence and Response defined by mode entropy. The dashed blue lines represent different isolines at different values of persistence-diversity function. And the red points represent initial experiments sampled. (C) Persistence vs. experiment iteration over the course of Bayesian Optimisation. Persistence domain, characteristic trajectory with different modes of motion highlighted, and temporal changes in the modes of motion over the course of the trajectory: (D-F) for trajectory with highest mode entropy, (G-I) for trajectory with highest persistence, (J-L) for trajectory with highest persistence-diversity function ($T$).

The second droplet ensemble is the other extreme case, with highest persistence and low diversity, giving target score, persistence and diversity of $T = 3115.78$, $P = 2.52 \times 10^5$, $S = 0.644$, respectively. The droplet composition is $\mathrm{ES} = 0.9$, $\mathrm{HDA} = 0.02$, $\mathrm{HCA} = 0.08$, $\mathrm{DEHPA} = 0$, and Figure 7G-I shows the persistence domain, characteristic trajectory, and modes of motion. The presence of a large amount of ES and lack of DEHPA makes the droplet move faster, as evident by the distribution of droplet states in the persistence domain. The characteristic trajectory ($\Delta\mathrm{t} = 530\mathrm{s}$) shows stable circular motion along the periphery of the Petri dish driven primarily by ES and no interactions with the base injections in the absence of DEHPA. This led to the droplet motion staying in global modes exclusively, maximising persistence.

The final droplet ensemble demonstrates the highest target score achieved over the entire experimental runs, towards the best trade-off between persistence and diversity, see Figure 7J-L. The droplet ensemble with oil ratios: $\mathrm{ES} = 0.67$, $\mathrm{HDA} = 0.03$, $\mathrm{HCA} = 0.17$, $\mathrm{DEHPA} = 0.13$ achieves $T = 2.6 \times 10^6$, $P = 2.64 \times 10^6$, $S = 1.367$. The persistence distribution, although lacking a density peak similar to the high persistence only, has a long-extended tail contributing to this ensemble's higher persistence compared to the highest diversity. The characteristic trajectory ($\Delta\mathrm{t} > 660\mathrm{s}$) shows both local and global modes showing droplet responding to the base injections and still sustaining higher persistence. The mode distribution for this trajectory shows a transition from global to local modes over the course of the experiment. This results in lower diversity than in the first case (highest diversity) but still much higher than in the second (highest persistence). Consequently, this represents the best trade-off mixture of the other two experiments, exhibiting more life-like behaviour as compared to other experiments. See Supplementary Media 4 for comparing three experimental videos with highest diversity, highest persistence, and highest persistence-diversity product. From these experiments, we conclude that DEHPA drives the strongest response of droplet motion to pH variations, in particular to the high pHs. To explore this observation, we performed additional experiments on the oil ratios with highest persistence-diversity product and its variant where DEHPA was replaced by ES. We observed a clear shift in droplet trajectories towards global motion indicating that droplets are less responsive to pH variations. See Supplementary Information Figures 15-17 and Supplementary Media 5.

## Conclusions

In this work, we explored the dynamics of self-propelled droplets through two complementary features of life-like behaviour: the ability to sustain activity over time and the capacity to diversify behaviour in response to internal or external change. By formalising these properties as persistence and diversity, we identified a trade-off between sustained activity and behavioural diversity within the systems examined and showed that droplet dynamics can be systematically tuned through chemical composition and environmental interactions. Mapping these dynamics onto a speed–shape complexity space provides a quantitative representation in which active and inactive states can be distinguished and persistence measured over time. Coupling these descriptors to autonomous experimentation enabled closed-loop robotic exploration of multicomponent chemical spaces and the identification of formulations with targeted dynamical behaviours. We investigated two oil-in-water droplet systems, incorporating programmable environmental perturbations through acid–base injections to explore persistence and persistence–diversity trade-offs. Together, these results provide a framework for engineering active chemical systems whose behaviour depends on sustained interactions between molecular composition, compartment dynamics and the surrounding environment.

An important next step is to extend this framework towards evolutionary droplet systems supported by more complex chemical metabolism. Incorporating reaction networks that consume chemical fuel, transform feedstocks and regenerate functional components could enable richer and more sustained behaviours. Controlling the supply of energy and matter would allow us to explore how these networks maintain activity further from equilibrium, and how their organisation influences persistence and responses to environmental change. A central challenge would be to couple this activity to the reproduction of molecular components and the re-establishment of functional networks in daughter or recipient droplets. When accompanied by heritable variation and differential replication, this would allow life-like behaviours to be investigated as traits of evolving droplet populations, see Figure 8. Assembly theory[1] provides a framework for investigating this extension by considering molecular assembly indices together with the copy numbers of the corresponding molecular objects. Applied to metabolically active droplets, it could connect the assembly complexity and abundance of molecular products to the networks responsible for their production and regeneration. Extending this analysis to network organisation would require physically meaningful definitions of the components and assembly operations, together with criteria for determining when a network has been reproduced. This would make it possible to ask whether complex chemical organisation can be reconstructed across successive droplets, increasing the copy number of that organisation while maintaining its constituent molecules through continued synthesis. Bringing assembly theory to bear on these systems would therefore connect measurable droplet behaviour with the construction, replication and selection of the chemical networks that sustain it. Combined with autonomous experimentation in reconfigurable environments, this approach could provide an experimental route from programmable active matter towards chemical systems capable of maintaining and propagating increasingly complex organisation.

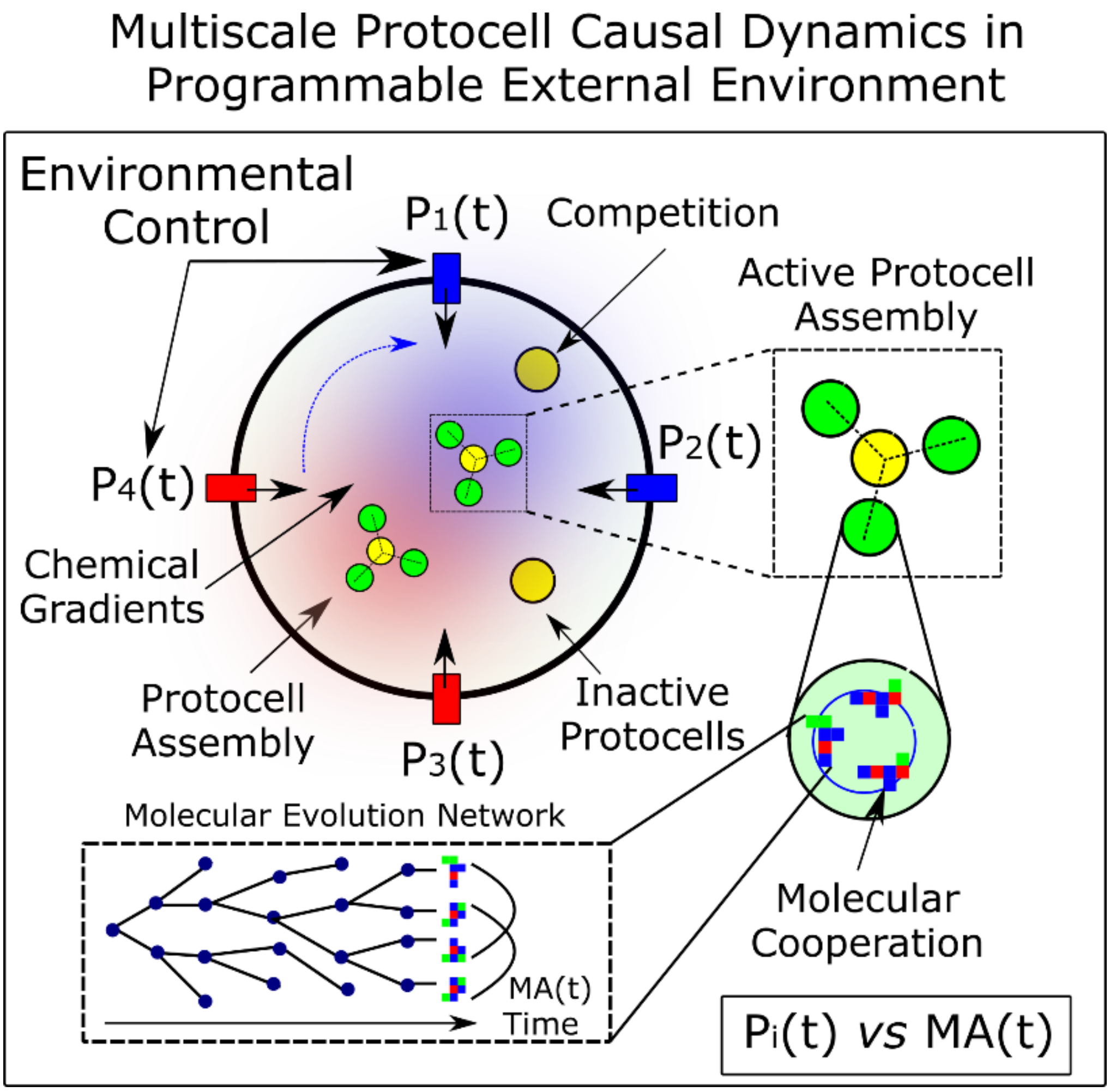

**Figure 8. Coupling the environment to molecular complexity.** Experimental Design for droplet-based Protocells Evolution using a robotic platform. The figure shows the experimental design comprising an arena with digital control to dynamically program environmental variables such as chemical gradients and external fluid flow and we will map the process variables $P_i(\mathrm{t})$ as a function of time and measure the emergence of molecular complexity as the assembly index increases.

# Methods

## Robotic Platform

We built an automated platform that mixed the oils to a given recipe, dispensed the droplets, recorded their motion, and cleaned the arena for the next experiment, see Figure 3A and B. Oils were transferred from stock, mixed and dispensed as droplets by an Opentrons OT-2 fitted with 300 µL and 20 µL single-channel pipettes. The 300 µL pipette carried out the oil transfers and the sampling of the aqueous phase, while the 20 µL pipette dispensed the droplets and transferred any oil whose share of the mixture was 20 µL or less. Mixing took place on a QInstruments BioShake 3000-T orbital shaker on the robot deck, which held the mixing vials in a custom rack of 60. The aqueous phase was dispensed, and the arena cleaned between experiments, by a custom 3D-printed unit driven by two independent TriContinent C-series syringe pumps, see Supplementary Information Figure 5. One pump served the aqueous line for cleaning water and surfactant solution, and the second served the waste lines, so that spent solution never shared a path with fresh solution. Acidic and basic solutions were injected through a 3D-printed holder that carried the Petri dish and four injectors, each with a port for acid and a port for base, eight injection points in total. The whole arena was recorded from above by a Raspberry Pi HQ Camera. Video analysis and the choice of the next recipe ran on a local PC, closing the loop (Figure 3C). One experiment occupied the platform for 40 to 60 minutes. Of this, 10 minutes went to mixing the oils and preparing the arena, and 2 minutes to dispensing the droplets. The video analysis and the scores took about a third of the recording time again. The platform ran without stopping, as the chemical stocks could be replenished and the waste emptied between the stages of an experiment. A full rack of mixing vials lasted at least two days of continuous operation. The platform is described in full in Supplementary Information Section 2.

## Chemical Systems

To create self-propelled oil-in-water droplets, we investigated two chemical systems. In each, droplets of an oil mixture moved through an aqueous phase: an ionic surfactant dissolved in DI water. In the first system, the aqueous phase was sodium dodecyl sulphate (SDS) at 100 mM**.** The oils were ethyl salicylate (ES), hydrocinnamaldehyde (HCA), bis(2-ethylhexyl) phosphate (DEHPA), and 2-hexyldecanoic acid (HDA). In the second system, the aqueous phase was tetradecyltrimethylammonium bromide (TTAB) at 20 mM, adjusted to pH 9.9–10.1 with NaOH. The oils were 1-octanol (Oct), octanoic acid (OA), diethyl phthalate (DEP), 1-pentanol (Pent), and DEHPA. From the oils a set of working stocks was prepared. Each stock was either a single oil or a fixed 80:20 premix of two oils. For each experiment, stocks were combined in the ratio given by the recipe, and droplets were dispensed from the resulting mixture. DEHPA, HDA and OA appeared only as the 20% component of a premix, so no droplet contained more than 20% of any of them. Above that fraction, oil mixtures did not produce self-propelled droplets reliably in preliminary tests. To make droplets visible, the stocks were dyed with Oil Red O at 12 mg per 45 mL. Full stock compositions are given in Supplementary Information Section 3. To create spatial and time-varying pH gradients in the aqueous phase, basic and acidic solutions were prepared. For the system with SDS, the basic solution was SDS (100 mM) with NaOH (100 mM), and the acidic solution was SDS (100 mM) with HCl (100 mM). For the system with TTAB, the basic solution was TTAB (20 mM) with NaOH to pH 12.9–13.1, and the acidic solution was TTAB (20 mM) with HCl to pH 0.9–1.1, see Supplementary Media 6 for examples of diverse droplet behaviour observed in in SDS and TTAB based aqueous phase.

## Experimental Protocol

Every experiment followed the same automated sequence, at ambient laboratory temperature, which was not recorded. The first experiments of a run used a predefined set of seed recipes, and every later recipe was proposed by the exploration algorithm. In the pure persistence run each seed recipe was run three times to confirm that the results were reproducible. No other experiment was repeated. Oils were combined according to the recipe into a 300 μL mixture in a mixing vial on the shaker and mixed for 10 s at 1000 rpm. The arena, a 6 cm Petri dish, was then rinsed four times with deionised water and refilled with 15 mL of fresh surfactant solution. The pipette aspirated 18 μL of the mixture and dispensed it as six droplets of 3 μL each onto the surface of the aqueous phase. The droplet volume was set to 3 μL as the greatest volume that stayed at the surface for every oil composition tested. Six droplets of that volume could fit into a single 20 μL pipette tip, so the arena could be loaded in a single aspiration. More droplets would have crowded the arena and drawn on the surfactant available in 20 mL of aqueous phase, and fewer would have interacted too rarely.

The droplets were recorded as a 730×730-pixel region of the sensor at 30 frames per second. Videos in pure persistence and persistence-diversity optimisation loops lasted for 17 minutes, while those for pH gradient effect experiments lasted for 30 minutes. In the experiments with induced pH gradient, the injections followed a fixed schedule. The first was made 60 s after the droplets were dispensed, and further injections followed every 180 s, cycling clockwise through the four injectors. In persistence-diversity experiments, the injection volume was 50 μL, while for pH gradient experiments it was 18 μL. After the recording the contents of the arena were transferred to waste and the protocol is described in full in Supplementary Information Section 2, see Supplementary Media 7 for droplet response to pH gradients with Thymol Blue dye in aqueous phase to observe high pH domains (left: 80% ES, 10% HDA, 10% DEHPA, right: 90% ES, 10% DEHPA).

## Video Analysis

Droplet motion was extracted from the recorded videos with software written in Python 3. OpenCV (version 4.13.0) was used for frame extraction and manipulation, with NumPy and SciPy used for general numerical analysis. The analysis proceeded in two stages. In the first, the software extracted the background image and the arena mask. To do this, the software selected 300 frames from the video: 100 from the beginning, 100 from the middle, 100 from the end. This set of frames was used to create the background image by calculating its inter-frame median. The background image was converted to greyscale and thresholded to produce the binary mask of the arena. To remove any artefacts from the mask, such as immobile droplets, the contour of the mask was traced, and every pixel within this contour was set to the foreground value. The background image and the mask were not reused across experiments, because the arena's position in the frame shifted by 1–2 pixels between runs. In the second stage, the software iterated over each frame of the video as follows. The background image from the first stage was subtracted from the frame using absolute difference. The binary mask was applied to the difference using bitwise AND operation on each of the colour channels. The masked difference frame was then converted to greyscale and thresholded to create the image of droplets' silhouettes. The silhouette image was decomposed into individual droplets using OpenCV's connectedComponentsWithStats function. Each component comprised information about the droplet: position, area in pixels, and the outline. Components with area ≤100 pixels were discarded. The outlines were used to calculate elliptic Fourier descriptors, from which the shape metric of each droplet was derived. Position, area, shape metric, and the descriptors for each droplet were recorded into a single feature table of the experiment. All subsequent analysis was performed using values from that table. A detailed description of the software is presented in Supplementary Information Section 2.

## Droplet Tracking

Droplet components in the feature table of the experiment were linked together to produce droplet trajectories. To achieve this, the components in the first frame were assigned unique indices, each marking the start of a trajectory. Each trajectory was continued into the nearest component of the next frame, provided that component lay within 20 pixels of the previous one. Trajectory continuations were resolved in the order in which components appeared in the feature table. A component could carry only one trajectory forward, so where two trajectories met at the same component - for example droplet collision - only the first trajectory continued and the second ended there. Any component that no trajectory was continued into started a new trajectory, which could happen even when the number of droplets was the same in both frames. A distance of 20 pixels between consecutive frames at 30 fps corresponds to a droplet speed of ≈ 61 mm.s$^{-1}$, which was never observed. The trajectory index of each component was recorded for further analysis of droplet metrics. The trajectory linking procedure is described in full in Supplementary Information Section 2.

## Droplet Metrics

Two metrics were derived for every droplet: its speed and a shape metric describing the complexity of its contour. Speed was derived per trajectory. The centre coordinates of a component along a trajectory were numerically differentiated with respect to the frame timestamps using NumPy's gradient function. The speed was calculated as the magnitude of the resulting velocity. Trajectories shorter than two frames were excluded, as a velocity cannot be defined for them. Speeds were computed in pixels per second and converted with the scale of the recordings, 0.1016 mm per pixel.

The shape metric was derived per component, in every frame in which it appeared. The contour of each component was decomposed into an elliptic Fourier series truncated at 10 harmonics, see Supplementary Information Section 1. Truncation smoothed the pixel-level steps of the extracted contour and set the finest contour feature the metric could resolve. The series was then evaluated at 300 evenly spaced points, at each of which the curvature of the contour was calculated analytically from the descriptors. The largest distance from the centre of the contour to its perimeter, $r_{\text{max}}$, was recorded at the same time. The bending energy of the contour was calculated as the integral of its squared curvature along the contour, divided by the square of the perimeter. The shape metric was defined as the base-10 logarithm of the product of that quantity and $r^3{}_{\text{max}}$. The two normalisations make the metric independent of droplet size: a circle gives $\log_{10}\left(\frac{1}{2\pi}\right) \approx -0.8$ regardless of its radius. The speed and the shape metric of every droplet in every frame were the two quantities from which the persistence score was computed. The definitions of both metrics and their parameters are given in full in Supplementary Information Section 1.

## Persistence and Diversity Scores

Two scores were calculated for each experiment: persistence, from the speed and the shape metric of every droplet, and diversity, from the modes of motion of the trajectories. The persistence score was calculated from a two-dimensional histogram of speed and shape metric. The speed and the shape metric of every droplet in every frame were collected into 500×500 bins, spanning 0 to 71 mm.s$^{-1}$ in speed and −1 to 6 in shape metric. Speeds and shape metrics outside these ranges were counted in the outermost bins. The score was the sum of the bin counts, each multiplied by a weight calculated with a kernel function, see Supplementary Information Section 1. The kernel set a threshold on each metric, 10 mm.s$^{-1}$ in speed and −0.5 in shape metric. A bin lay in the living domain when both of its metric values were above these thresholds, and in the non-living domain otherwise. Above the thresholds the kernel grew only logarithmically, and below them it decayed exponentially. We chose this shape to favour oil recipes whose droplets spent most of the experiment in the living domain over recipes with isolated extremes of speed or shape. The score was a weighted count and not an average, so it also grew with the number of droplet-frames in an experiment. Diversity was calculated from the trajectories longer than three frames. Each frame of a trajectory was classified along two dimensions. The first dimension separated global from local motion. The positions of the droplet within a sliding window of 151 frames, or 5 s, were enclosed by their convex hull. The motion was global when the diameter of the hull exceeded 13.7 mm, a quarter of the arena diameter.

The second dimension separated turning from straight motion. The positions of the droplet were smoothed with a fourth-order Butterworth low-pass filter at a cutoff frequency of 4 Hz. The filter was applied forwards and backwards with SciPy's sosfiltfilt function, so it introduced no phase shift. From the filtered positions, the velocity and the acceleration were obtained by numerical differentiation. The motion was turning when the angle between them was within 45° of a right angle. The two dimensions gave four modes of motion, and the diversity score was the Shannon entropy of the counts of these modes over all trajectories of the experiment. The entropy was calculated with natural logarithms, so the score ran from 0 for a single mode to ln 4, or 1.39, for four modes in equal proportion. Unlike persistence, the entropy depended only on the proportions of the modes and not on the number of droplet-frames. The diversity score is defined in full in Supplementary Information Section 1.

## Modelling and Exploration Algorithm

New oil recipes were proposed from a model built on the recipes and target scores collected so far. In the pure persistence run the target score was the persistence score. In the combined persistence–diversity run it was the persistence score multiplied by the tenth power of the diversity score. After every experiment the target score was modelled as a function of the oil ratios in the recipe. The model was scikit-learn's GaussianProcessRegressor, with a Matérn kernel added to a WhiteKernel that absorbed the experimental noise. The smoothness parameter of the Matérn kernel was fixed at $\nu = 1/2$, which assumes that the target score can change abruptly between neighbouring recipes. Oil ratios were standardised with scikit-learn's StandardScaler before they entered the model, and the scores were centred and scaled to unit variance by the regressor. Three strategies took turns in a fixed rotation, each with its own criterion for choosing the next recipe. The rotation was fixed so that half of the experiments exploited the model and half explored. In the exploiting experiments the criterion was the predicted target score, and the recipe with the highest value was taken. The exploring experiments alternated between two criteria. One was the Euclidean distance to the prior dataset, and the recipe at the greatest distance was taken, filling gaps in the space of oil recipes. The other was the gap between the scores predicted so far, and the recipe whose predicted score fell in the widest gap was taken, filling gaps in the range of scores. Every strategy searched the space of oil recipes in the same way. First, 1000 candidate recipes were drawn from a Dirichlet distribution, which samples the recipes uniformly and gives ratios that sum to one by construction. The criterion was then evaluated for every candidate. The best candidate became the starting point of a local optimisation with SciPy's minimize function, applied to the criterion and negated where the criterion was to be maximised. The refined recipe was kept only when it improved on that candidate. Otherwise, the best sampled candidate was submitted as the next recipe. Modelling and recipe suggestion are described in Supplementary Information Section 4.

## NMR Measurements

Oil concentrations in the aqueous phase were measured by $^1$H NMR in separate experiments. These repeated the reported recipes under the same conditions, except that no video was recorded and the OT-2 pipette sampled the arena. At fixed intervals the pipette withdrew a 300 µL aliquot of the aqueous phase into a vial. Each aliquot was diluted with 300 µL of solution before measurement. For the SDS system this was 0.1 M TFA in water, which holds the solutes in their protonated forms. For the TTAB system it was the basic injection solution, TTAB (20 mM) with NaOH, which holds them deprotonated. Spectra were recorded on a Bruker 600 MHz spectrometer at 298 K, with water suppression by WATERGATE with pre-saturation. Reporter signals were assigned by comparison with spectra of the pure oils in the same surfactant solution. Concentrations were obtained by integrating the reporter signals in MestReNova, scaled against the surfactant signal. The amount of each oil remaining in the droplets was not measured directly, but obtained by mass balance, as the difference between the amount dispensed and the amount in the aqueous phase. More details are provided in full in Supplementary Information Section 5.

## Author Contributions

L.C. generated the original idea, and the L.C. and A.S. conceived the idea and designed the research plan. A.S. developed the theoretical framework with input from I.S. A.S. implemented the out-of-loop analysis in Mathematica, and I.S. implemented the in-loop analysis and the Bayesian optimisation in Python. M.J. designed the OT-2 module for the Petri dish and the 3D-printed surfactant dispensing unit, and I.S. designed the Petri dish holder with the acid and base injectors. I.S. and M.J. wrote the control software for the platform. A.S. and K.L. performed the preliminary experiments, and I.S. performed the exploration experiments on the platform and managed the data. M.J. performed the NMR analysis and calculated the dissolution rates with inputs from K.L. A.S., I.S., A.S. and L.C. wrote the paper with help from the rest of the authors. I.S. and A.S. contributed equally to this work.

## Data Availability

The characteristic source data which include experimental and simulated data used for generating the figures in the main text and supplementary information are available on Zenodo. Additional data due to large size are available upon request to the corresponding author at Lee.Cronin@glasgow.ac.uk

## Code Availability

All the code required for running experiments, algorithms, and data analysis is available on https://github.com/croningp/droplet_persistence.

## Acknowledgements

We acknowledge financial support from the John Templeton Foundation (grant nos. 61184 and 62231), the Engineering and Physical Sciences Research Council (EPSRC) (grant nos. EP/L023652/1, EP/R01308X/1, EP/S019472/1 and EP/P00153X/1), the Breakthrough Prize Foundation and NASA (Agnostic Biosignatures award no. 80NSSC18K1140).